\documentclass[a4paper, 10pt, conference]{ieeeconf}
\usepackage{graphicx}
\usepackage{ifthen}
\usepackage{amssymb}
\usepackage{listings}
\usepackage{color}
\usepackage{amsmath}
\usepackage{listings}
\usepackage{array}

\usepackage[ruled,lined,algonl,boxed]{algorithm2e}
\usepackage[para]{footmisc}
\usepackage[noend]{algpseudocode}

\title{A Top-Down Approach to Managing Variability in Robotics Algorithms}

\author{\authorblockN{Selma Kchir, Tewfik Ziadi, Mikal Ziane}
\authorblockA{\\UMR CNRS 7606 LIP6-MoVe\\
       Universit\'e Pierre et Marie Curie, France\\
       Email: firstname.lastname@lip6.fr}
\and
\authorblockN{Serge Stinckwich}
\authorblockA{\\UMI UMMISCO 209 IRD/UPMC\\
		Universit\'e de Caen-Basse Normandie, France\\
		Email: serge.stinckwich@ird.fr}
}

\begin{document}
\maketitle
\begin{abstract}
One of the defining features of the field of robotics is its breadth and heterogeneity. Unfortunately, despite the availability of several robotics middleware services, robotics software still fails to smoothly handle at least two kinds of variability: algorithmic variability and lower-level variability. The consequence is that implementations of algorithms are hard to understand and impacted by changes to lower-level details such as the choice or configuration of sensors or actuators. Moreover, when several algorithms or algorithmic variants are available it is difficult to compare and combine them.

In order to alleviate these problems we propose a top-down approach to express and implement robotics algorithms and families of algorithms so that they are both less dependent on lower-level details and easier to understand and combine. This approach goes top-down from the algorithms and shields them from lower-level details by introducing very high level abstractions atop the intermediate abstractions of robotics middleware. This approach is illustrated on 7 variants of the Bug family that were implemented using both laser and infra-red sensors.
\end{abstract}
\section{Introduction}
The development of robotics software must deal with a large amount of variability from at least two sources. First, robots are very different from each other: "They have different locomotion mechanisms, different onboard computational hardware, different sensor systems, and different sizes and shapes." \cite{Smart07}. Second, robots are used for very different tasks with varying constraints which leads to a large variety of algorithms.

Robotics middleware, among other improvements, has been a significant attempt at addressing the first kind of variability by decoupling robotics application from lower levels details. "It is designed to manage the heterogeneity of the hardware, [...] simplify software design [...]. A developer needs only to build the logic or algorithm as a component" \cite{Elkady12}.

The task is huge however, as well explained by W.D. Smart who considers the hypothetical case of middleware providing an obstacle-avoidance routine for a mobile robot \cite{Smart07}. According to him "writing a generic
obstacle avoider that will work for all locomotion mechanisms, using input from all possible sensors is a daunting task". It is not surprising then that, a few years later, middleware services are still far from solving this decoupling problem.

Consequently, robotics algorithms are still
\begin{itemize}
\item difficult to understand,
\item difficult to adapt or combine,
\item impacted by changes in lower-level details.
\end{itemize}

Even with the very simplified assumptions such as those of the Bug family of navigation algorithms \cite{Lumelsky84} it is far from obvious in which case such or such variant is best suited to go from one point to another while avoiding obstacles \cite{JNG07}.
In this paper we propose a top-down approach to complement the bottom-up middleware approach. The input of this approach is a robotic task and either a family of algorithms or at least enough knowledge to produce algorithms to solve it. Its output is twofold:
\begin{itemize}
\item a set of algorithmic, sensory or action abstractions,
\item a configurable generic algorithm.
\end{itemize}

The produced algorithm can be configured by providing the values of a series of parameters to be adapted to different hypotheses, say on the environment. Furthermore it is generic in two other ways. First, it is decoupled from low-level details on sensors and actuators and second, the algorithmic variability which cannot be resolved statically by specifying configuration parameters is managed by dynamically linking the actions abstractions to executable routines.

This rest of this paper is organized as follows. Section II describes our approach and illustrates it on Bug algorithms. Section III deals with the problem of organizing reusable implementations of the abstractions. Section IV discusses the approach. Section V presents preliminary validation results while section VI compares our approach to related work.

\section{Rational}
In this section, we propose our approach to provide a generic algorithm for a family of algorithms for mobile robots. All the algorithms of the family are assumed to achieve the same task.
It is further assumed that there is a significant plan aspect in the algorithms: they are not merely reactive.
The generic algorithm can then be seen as a planner specialized to the considered task. 

Our approach is to be performed by a human robotics expert with a strong background in programming.
The input to the approach is a description of the task to be accomplished and a series of algorithmic descriptions in whatever format the expert understands (code, pseudo-code ...).
The approach consists in defining a generic algorithm while extracting two kinds of abstractions to shield it from respectively low-level and algorithmic sources of variability. These two steps are performed concurrently. In a third subsequent step, the generic algorithm is implemented using the Template Method design pattern which delegates (similarly to the Bridge design pattern) sensing and actuation to so-called virtual sensors and actuators whose implementations can then vary independently from the algorithmic variants.

\begin{figure}[!htb]
\centering
\hspace{-0.5cm}
\includegraphics[scale=0.28]{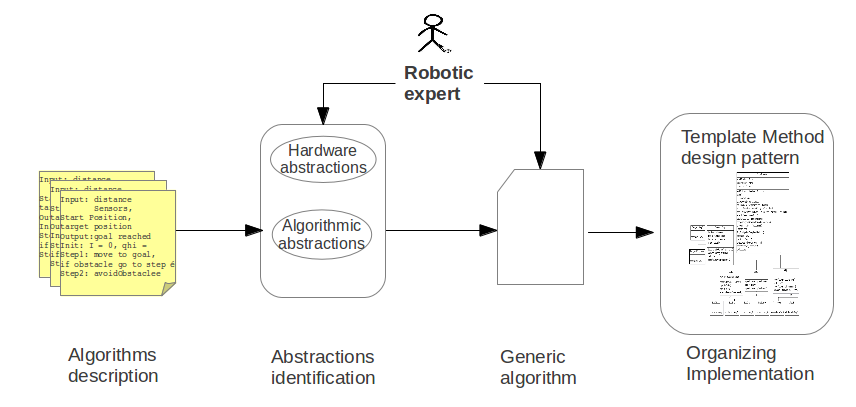}
\caption{Generic algorithm: from identification to implementation}
\label{approach}
\end{figure}

 Figure \ref{approach} depicts the main steps of our approach (the first two steps are preformed concurrently):
\begin{enumerate}
 \item \textbf{Abstractions Identification}
\begin{itemize}
\item Definition: Abstractions are a set of methods which allows to encapsulate specific data or to abstract specific methods by extracting a general signature from specific ones.
\item Description: If we consider a family of algorithms where variants propose
different strategies to perform the same task, we should handle this variability. Therefore, abstractions must gather hardware and what we call algorithmic variability.
One of the possible approaches to define abstractions is a goal directed approach.
In our point of view, defining hardware abstractions does not require specific knowledge of specific robots sensors or actuators but a global description of the needed data types of the handled task.
Regarding algorithmic abstractions they relie on a comprehensive study of algorithms variants whence the needed intervention of a robotic expert. Invariant parts among algorithms of the same
family can then be written in terms of these abstractions.
\end{itemize}
\item \textbf{Generic Algorithm Definition}
\begin{itemize}
\item Definition: A generic algorithm is a sequence of instructions written in terms of hardware and algorithmic abstractions.
\item Decription: At this level, the robotic expert combine the abstractions identified previsouly in order to write the algorithm or to design a state machine corresponding to a specific robotic task.
In this way, there is no specific detail related to a particular algorithms variant and the defined algorithm is completely independent from low level details and from specific algorithms variants.
\end{itemize}
\item \textbf{Organising Implementation}
\begin{itemize}
\item Definition: Organising implementation means implementing variants of a family of algorithms starting from the generic algorithm.
\item Description: In this step, we implement the invariant parts of the algorithm in terms of abstractions. To do so, the Template Method (TM) design pattern \cite{Gamma95} allows the definition of
an algorithm skeleton in an abstract class as a template method.
This method is the generic algorithm identified previsouly. It defines the basic step of the algorithm as "placeholders" methods (or hook methods), that are different in each subclass.
Invariants parts of the algorithm are represented
once in the abstract class as concrete methods. Abstract methods will be used to represent needed operations that are different in each subclass.
The main idea is to represent the variants of the algorithm as subclasses that implement these abstract methods. Thus, algorithmic abstractions will be implemented in these subclasses.
Concerning hardware abstractions, they are not implemented in subclasses like algorithmic abstractions but delegated to adaptors which implement them. We define one adapter per physical
sensor. Adaptors implement these operations to extract physical data and convert them to the needed abstractions. Then, the specification of the physical sensor used in the algorithm is done at the deployment level.

\end{itemize}
\end{enumerate}

In the next section, we illustrate our approach on the Bug family of navigation algorithms. 
\section{Case study: Bug algorithms}
It is symptomatic that even with the very simplified assumptions underlying these algorithms it is still difficult:
\begin{itemize}
\item to understand the differences among the algorithmic variants,
\item to choose one variant over another,
\item to share development efforts among variants.
\end{itemize} 

\subsection{Bug algorithms overview}
The Bug algorithms attempt to solve the navigation problem in an unknown two-dimensional environment with fixed obstacles and a known goal position. Over 19 versions of 
Bug algorithms have been defined in the literature. Among them, 7 variants are considered in this paper.

Bug algorithms share the following assumptions~\cite{Lumelsky84} : 1) The environment is unknown and a finite number of fixed obstacles are placed arbitrarily.
2) The robot is considered as a point (i.e. without body). It has perfect sensors (for obstacle detection) and a perfect localization ability (e.g. to compute its distance to its goal).

In Bug family, the robot has a local knowledge and a global goal. In other words, the inputs of Bug algorithms are the robot's start position and the target position.
At the end of the algorithm's execution, the robot must indicate if its goal is reached or if the goal is unreachable.
Most of the Bug algorithms can be programmed on any mobile robot using tactile or distance sensing  and a localization method while some require distance sensing.

\subsection{Abstractions identification}
%


In the litterature, as far as we know, Bug algorithms are only published with an informal description.
Bug algorithms are very similar fundamentally but differ in some points of interest. Their principle is the following: (1) The robot \textbf{motion to its goal} until an \textbf{obstacle is detected}
 on its way. (2) From the point where the obstacle were encountered (called hit point), the robot \textbf{looks for a point} (called leave point) \textbf{around the encountered obstacle} to be able to move to its goal again. 
(3) When a \textbf{leave point is identified}, the robot \textbf{moves to it} and leaves the obstacle. 
Steps (1), (2) and (3) are repeated until \textbf{the goal is reached} or until the robot indicates that the \textbf{goal is unreachable}.
This description of Bug hides hardware and algorithmic variability. Hardware variability impacts on obstacle detection and localisation tasks because of their dependency of the robot actual sensors.
Algorithmic variability is related to the leave point identification task.
Following a goal directed approach and based on a comprehensive study of Bug algorithms, we have identified the following abstractions (described here as methods):

\begin{itemize}
\item Motion to goal: the robot needs to face its goal and to be able to go ahead it: {\sc faceGoal(Point goalPosition)}, {\sc goAhead()}.
\item Obstacle on the robot's way: {\sc bool obstacleInFrontOfTheRobot()}
\item Current position: {\sc Point getPosition()}
\item Get around the encountered obstacle consists in performing clockwise or anticlockwise circumnavigation while maintaining a safety distance from the obstacle: 
{\sc wallFollowing(direction)}
\begin{itemize}
\item {\sc double getSafeDistance()}
\item {\sc bool obstacleOnTheRight(), bool obstacleOnTheLeft()}: Depending if we follow the wall on the right or the left.
\item {\sc double getRightDistance(), double getLeftDistance()}: To keep a safety distance from a wall, we need to know the distance to the wall from the right and the distance from the left.
\end{itemize}
\item Look for a leave point: {\sc identifyLeavePoint(bool direction, Point robotPosition, Point goalPosition)} which consists in wall following while looking for a leave point.
This could be a local decision (choose the first leave point which satisfies the algorithm condition) or a global decision (choose the best leave point after visiting all the points
around the obstacle). The following conditions are examples of decisions that the algorithm defines to find a leave point:
\begin{itemize}
\item \textbf{The closest point around obstacle boundary condition.} It consists on recording the closest point to the goal among all the points ever visited by the robot while performing boundary following\cite{Lumelsky84}\cite{Lumelsky86}.
\item \textbf{The m-line detection condition.} It is a straight line between the starting point and the target which aims at providing a set of prededefined leave points. The robot must leave only on these points\cite{Lumelsky86}\cite{Noborio00}.
\item \textbf{The local minimum condition.} Using its distance sensors, the robot can detect discontinuity points\cite{Kamon98} on obstacles on its way with respect to the target.
\item \textbf{The disabling segment condition.} A disabling segment occurs when the robot cannot move to its goal from all points in a segment while perfoming boundary following.
\item \textbf{The step method.} The robot uses its distance sensors to detect a point which is \textit{STEP}\cite{Kamon95} closer to the target than any point already visited.
\end{itemize}
Thus, we have identified the method {\sc findLeavePoint(Point robotPosition, Point hitPoint, Point goalPosition)} which will be applied to each point around the obstacle 
and which is specific to each algorithm variant to hide all low level details. The code of {\sc identifyLeavePoint} is given by the algorithm \ref{identifyLP}.
\item Leave point identified: {\sc bool isLeavePointFound()}
\item {\sc researchComplete}: In case of a local decision of leave point identification, the research complete is equivalent to the condition leave point identified. In case of a global decision, 
the research complete decision is equivalent to a complete cycle around the obstacle: {\sc researchComplete(Point robotPosition, Point hitPoint, Point goalPosition)}.
\item Move to the leave point: Once the leave point identified, the robot goes to it. This is a variability point because the robot can go to the leave point following the shorter distance or other 
strategies: {\sc goToLeavePoint(Point leavePoint)}.
\item {\sc goal unreachable}: This condition is checked if there is no leave point identified after performing an entire cycle around the obstacle: {\sc bool completeCycleAroundObstacle(Point robotPosition,Point  hitPoint) and not isLeavePointFound()}
\item goal reached: Depending on the algorithm objective, we define an error margin which indicates if the robot must reach its goal or stops before arriving to it: {\sc bool goalReached(Point robotPosition, Point goalPosition, double err)}
\end{itemize}

\begin{tiny}
\begin{algorithm}
\hspace{-1cm}
\label{identifyLP}
\caption{Identify leave method algorithm}
\begin{algorithmic}
    \Function{identifyLeavePoint}{$Bool \ direction,\ Point \ robotPos,\ Point\ goalPos$}{\\
       \BlankLine
 	computeData(robotPos);\\
	wallFollowing(direction);\\
        findLeavePoint(robotPos, hitPoint);\\
 }
 \EndFunction
\end{algorithmic}
\end{algorithm}
\end{tiny}
After identifying these abstractions, we classify them into hardware abstractions and algorithmic abstractions.
\begin{enumerate}
 \item Hardware abstractions:

{\sc getPosition()}, {\sc getSafeDistance()}, {\sc obstacleOnTheLeft()}, {\sc obstacleOnTheRight()}, {\sc getRightDistance()}, {\sc getLeftDistance()}, {\sc obstacleInFrontOfTheRobot()},
 \item Algorithmic abstractions:

{\sc findLeavePoint(Point robotPosition, Point hitPoint, Point goalPosition)}, {\sc identifyLeavePoint(bool direction, Point robotPosition, Point goalPosition)}, {\sc bool goalReached(Point robotPosition, Point goalPosition, double err)},
{\sc completeCycleAroundObstacle(Point robotPosition,Point hitPoint)}, {\sc bool isLeavePointFound()}, {\sc goToLeavePoint(Point leavePoint)}.
\end{enumerate}

\subsection{Generic Bug algorithm definition}

The generic algorithm is a combination of the previously defined abstractions as a sequence of instructions. Our generic algorithm is given in \ref{algoGen}.

\begin{tiny}
\begin{algorithm}
\hspace{-1cm}
\label{algoGen}
\caption{Bug generic algorithm}
\SetKwInOut{Input}{input}
\SetKwInOut{Output}{output}
\SetKwData{start}{$q_{start}$}
\SetKwData{target}{$q_{goal}$}
\SetKwData{distance}{$\varphi(p, \theta)$} 
\SetKwData{mline}{$M-line$}
\SetKwData{minDistGoal}{$minDist2Goal$} 
\SetKwData{pathToMin}{$path2leavePoint$} 
\SetKwData{total}{$totalPath$} 
\SetKwData{iterator}{$i$}
\SetKwData{hitPoint}{$q_H^i$} 
\SetKwData{leavePoint}{$q_L^i$}
\SetKwData{hlist}{$H_{list}$}
\SetKwData{llist}{$L_{list}$}
\SetKwData{position}{$x$}

\SetKwFunction{MotionToGoal}{motionToGoal()}
\SetKwFunction{termination}{EXIT\_SUCCESS}
\SetKwFunction{FCGoal}{faceGoal()}
\SetKwFunction{wall}{wallFollowing}
\SetKwFunction{Stop}{stop()}
\SetKwFunction{GoTo}{goToLeavePoint}
\SetKwFunction{idLP}{identifyLeavePoint}
\SetKwFunction{lpFound}{leavePointFound}
\SetKwFunction{RC}{researchComplete}
\SetKwFunction{cycle}{completeCycleAroundObstacle}
\SetKwFunction{obstacleDetected}{obstacleInFrontOfTheRobot()}
\SetKwFunction{loc}{getPosition()}
\SetKwFunction{Reachable}{goalIsReachable}
\SetKwFunction{TReached}{goalReached}
\SetKwFunction{failure}{EXIT\_FAILURE}
\SetKwInOut{sensor}{Sensors}
\SetKwInOut{init}{Initialisation}
\SetKwFunction{dir}{getDirection()}


\SetKwFunction{rec}{computeData}
\sensor{A perfect localization method. \\ An obstacle detection sensor}
\Input{Position of Start (\start), Position of Target (\target)}
\init{robotPos $\gets$ \loc; direction $\gets$ \dir;}
\BlankLine
\If{\TReached{robotPos}} {\termination; }

\ElseIf{\obstacleDetected == true} {
\idLP(direction, robotPos, goalPos)\;
\If{\lpFound{} $\&\&$ \RC{robotPos, getHitPoint(), goalPosition}}{
\GoTo{getLeavePoint()}\;
\FCGoal{}\;
}
\ElseIf{\cycle{robotPos, getHitPoint()} $\&\&$ !\lpFound{}} {\failure;}
}

\Else{\MotionToGoal;}\

\end{algorithm}
\end{tiny}

\subsection{Implementation: Template Method design pattern}

\begin{figure*}[!htb]
 \hspace{-1cm}
 \centering
 \includegraphics[scale=0.3]{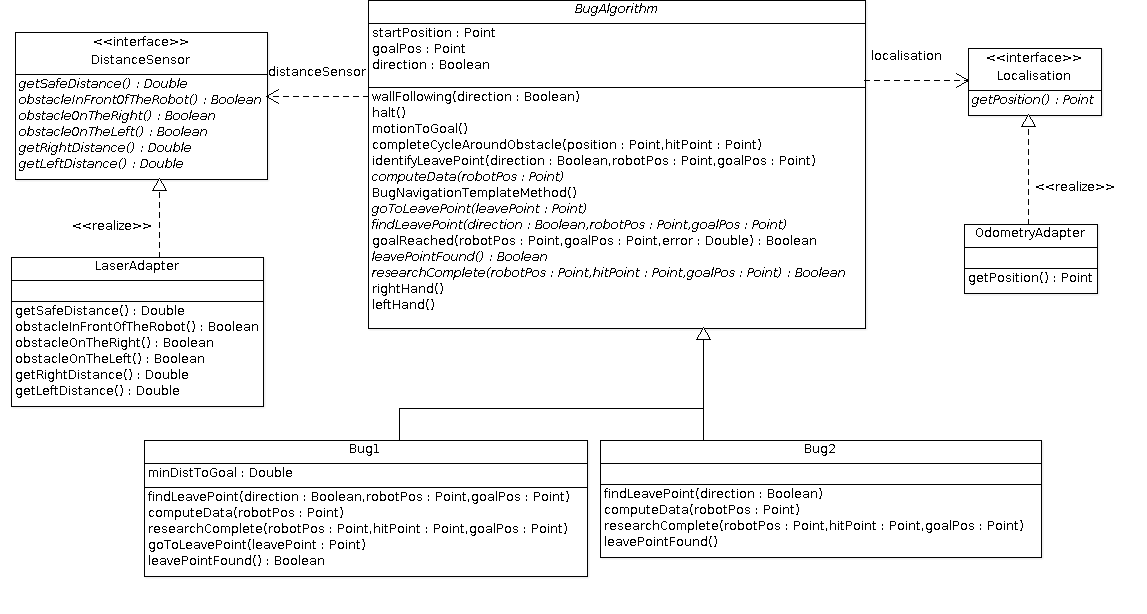}
 \caption{Bug algorithms class diagram (extract)}
 \label{Dclass}
 \end{figure*}

As we said previously, the Template Method deals with variability problem by proposing to define the basic step of the algorithm as a template method written in terms of abstractions which will be 
implemented in subclasses. Define a subclass for each algorithm without handling sensors and actuators variability cause a combinatorial explosion. Consequently, we dealed with this problem by 
delegating hardware variability to what we call virtual sensors or adaptors. Virtual sensors convert specific data from the physical sensors to the needed data in the algorithm.
They implement a set of interface abstract operations. For instance, the method {\sc obstacleInfrontOfTheRobot()} is different in each sensor adaptor.
We have defined an adaptor per sensor. In case of multiple physical sensors use, their adaptors are combined together to provide the needed data for the algorithm.
The architecture of our implementation is presented in Figure \ref{Dclass}.

To perform wall following, right hand and left hand algorithms are written in terms of sensors abstractions (i.e. {\sc obstacleOnTheLeft()} and {\sc obstacleOnTheRight()})
\subsection{Discussion}
The hierarchy of classes defined by our approach can lead to a combinatorial explosion if we add additional variants of Bug family. It is then much cheaper to build operations of the 
algorithm as components and to assemble desired family members from them.
This is our main motivation for using an alternate solution based on Software Product Lines (SPL).
SPL are easy to use, compact and generate only a configuration which interests the user.
This constitutes an immediate topic for a future work.

\section{Results and Validation}

Several performance comparison studies\cite{JNG07} were realized on the Bug family. In this section, we do not intend to perform a comparison between Bug family but to prove that each 
algorithm of the Bug family fits with our generic algorithm.\\
We demonstrate the capabilities of our generic algorithm in the OROCOS-RTT framework through 7 variants of Bug: Bug1~\cite{Lumelsky84}, Bug2~\cite{Lumelsky86}, Alg1~\cite{Noborio00}, Alg2~\cite{Sankaranarayanar90:Alg2}, DistBug~\cite{Kamon95},TangentBug~\cite{Kamon98} and Rev1~\cite{Horiuchi01}. 

\begin{figure}[!htb]
\label{simulation}
\begin{center} 
\begin{minipage}[b]{.46\linewidth}
\includegraphics[scale=0.16]{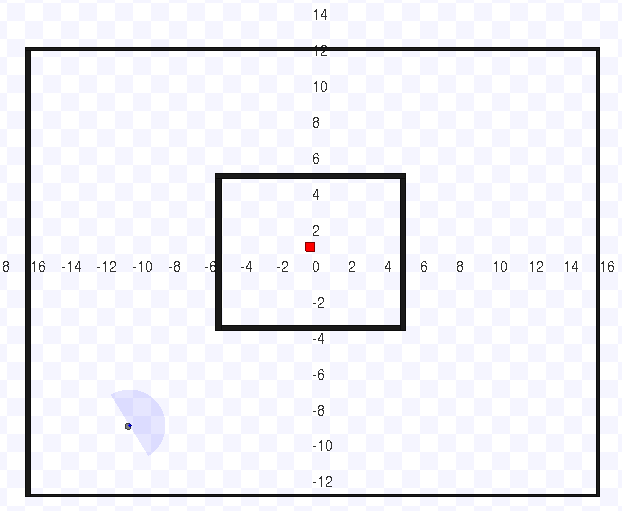} 
\caption{Environment with unreachable target}
\label{env01}
\vspace{-0.3cm}
\end{minipage}
\begin{minipage}[b]{.46\linewidth}
\vspace{-0.3cm}
\includegraphics[scale=0.15]{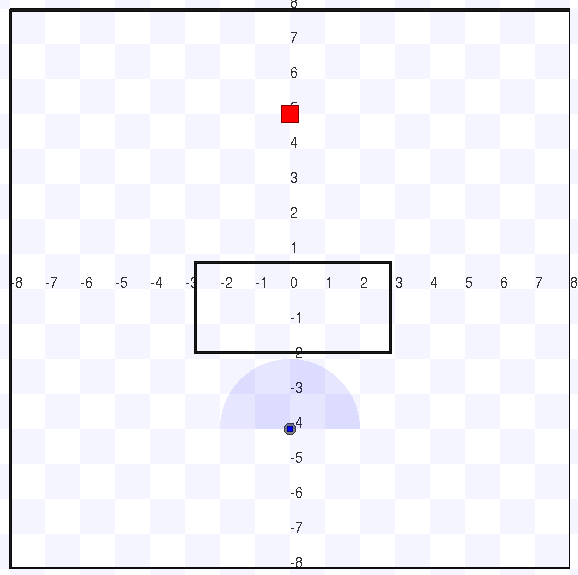} 
\caption{Basic Simulation Environment}
\label{env2}
\end{minipage}
\begin{minipage}[b]{.46\linewidth}
\includegraphics[scale=0.2]{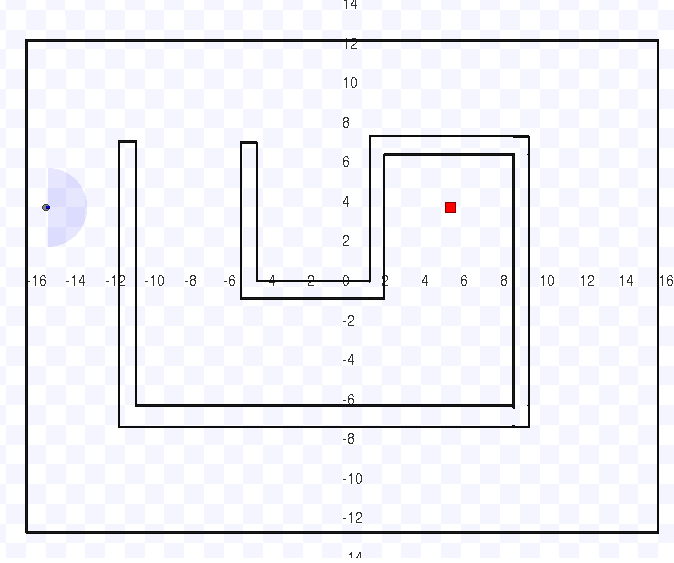}
\caption{Environment to validate the encountered points condition}
\label{env3}
\vspace{-0.3cm} 
\end{minipage}\hfill
\end{center} 
\end{figure}

Simulation was performed using Stage-ROS with 2 configurations. \\
The first tested configuration was done with a laser scanner with 180 degrees scanning angle and a detection range which varies from 0.02 meters to approximately 4 meters and a GPS for localization. The robot does not have any knowledge about its environment except its start position and the goal position.\\
The second configuration was done using 3 infrared range sensors placed on the front, on the left and on the right compared to the central axis of the robot with a field of view equals to 26 degrees and a range which detects until 2 meters. \\
To demonstrate that any algorithm of the ones studied here fits with our generic algorithm, we have tested both configurations in 3 different environments for all algorithms. \\ 
We defined the first environment (see figure \ref{env01}) to validate the target reachability condition.
In all implemented algorithms, the robot returns failure because it can not achieve its goal.
The second environment shown in figure \ref{env2} is a simple simulation environment with one obstacle. Some algorithms behave similarly in this environment despite their different 
obstacle avoidance strategies. For instance, Alg1 and Bug2 rely on the M-line detection condition but behave differently when they encounter an already traversed point. In other words, 
Alg1 was defined above Bug2 to overcome situations where the robot can find itself in an overlasting loop around obstacle. For this reason, we defined the third environment, presented in figure \ref{env3}, to validate the encountered points condition (i.e. when the robot encounters an already traversed point), particularly used in the algorithms Alg1 and Alg2. \\

The trajectory of the robot could depend on the computation time of sensors information.
Since most of the algorithms treatments are reactive, we had to define the period of execution time which allow us to get as much real-time information as possible and to optimize the execution time and the robot's path.
Consequently, we set the execution period of the algorithm to 0.5 seconds.


Alll the code of Bugs algorithm variants\footnote{https://github.com/SelmaKchir/BugAlgorithms} and simulations results\footnote{https://github.com/SelmaKchir/BugAlgorithms/wiki/Implementing-Bug\-Algorithms-variants} are available on GitHub.

\section{Related work}

Several software engineering technologies and methods aim at improving software design and reusability. In robotics, reusability is obtained after handling variability which could be related 
to the robot hardware or to the robot's capabilities (e.g. different strategies for obstacle avoidance, etc).
Robotics software like Robot Operating System (ROS)\cite{Quigley09}, RISCWare framework \cite{Elkady13}, Player \cite{Collett05} propose a hardware abstraction layer to encapsulate the sensors 
that gather information about the environment and provide a set of predefined intefaces. These middlewares rely on a bottom up approach which consists on classifying the most used physical 
sensors (or actuators), analysing the potential data they are able to provide and then define interfaces. \\
Unlike these middlewares, we use a top-down approach which consists in analysing the needed data in a particular application, defining abstractions and then writing the application in terms of 
these abstractions (possibly provided by middlewares).

In our approach, we have applied the Template Method design pattern to implement variants of a generic algorithm.
They are several proposals that try to apply design patterns in robotics to handle variabilities. CLARATy authors \cite{Nesnas:2003fk} say they used many well-know techniques developed by the software community including design patterns but without being much more explicit about them.
MARIE \cite{Carle06}, a middleware framework for robotics, applied the Mediator Design Pattern \cite{Gamma95} to create a mediator interoperability layer for distributed robotics applications.\\

\section{Conclusion and Perspectives}
In this paper we have proposed an approach to organize families of algorithms so that the algorithmic decisions to make are clearly expressed and decoupled from implementation details. 
Our approach relies on the Template Method design pattern to define a generic algorithm for Bug algorithms family.

Our approach consists of three steps.
The first step takes as input a set of algorithmic variants; as described in the litterature, and manually extract hardware abstractions and what we have called algorithmic abstractions.
The generic algorithm is then defined as a sequence of instructions in terms of these abstractions.
The implementation of these variants relies then on the Template Method design pattern.

This approach has been illustrated on the Bug family of robot navigation algorithms. Seven implemented variants of Bug algorithms have been implemented 
using the OROCOS-RTT robotic framework. Simulation was performed in different unknown environments with a random positioning of obstacles.

Bug variants are about 20 versions of algorithms. Implementing all \texttt{BugAlgorithm} subclasses is very complicated to understand and too expensive to build all Bug family members. 
In addition, it is error prone to not define constraints on subclasses and the data they must specify and use.\\
It is then much cheaper to build operations of the algorithm as components and to assemble desired family members from them. 
This is our main motivation for using Software Product Lines (SPL)~\cite{Clements01} as an immediate topic of future work.




\bibliographystyle{IEEEtran}
\bibliography{IEEEabrv,main}

\end{document}